\documentclass[sigconf]{acmart}
\usepackage{bm}
\usepackage{multirow}

\usepackage{balance}

\setcopyright{acmcopyright}

\begin{document}

\copyrightyear{2019}
\acmYear{2019}
\acmConference[MM '19]{Proceedings of the 27th ACM International Conference on Multimedia}{October 21--25, 2019}{Nice, France}
\acmBooktitle{Proceedings of the 27th ACM International Conference on Multimedia (MM '19), October 21--25, 2019, Nice, France}
\acmPrice{15.00}
\acmDOI{10.1145/3343031.3350978}
\acmISBN{978-1-4503-6889-6/19/10}

\fancyhead{}

\title{Long Short-Term Relation Networks for Video Action Detection}

\author{Dong Li, Ting Yao, Zhaofan Qiu, Houqiang Li and Tao Mei}
\affiliation{University of Science and Technology of China, Hefei, China \and JD AI Research, Beijing, China}
\email{{{dongli1995.ustc, tingyao.ustc, zhaofanqiu}@gmail.com;lihq@ustc.edu.cn;tmei@jd.com}}

%
\renewcommand{\shortauthors}{Li and Yao, et al.}

%
\begin{abstract}
  It has been well recognized that modeling human-object or object-object relations would be helpful for detection task. Nevertheless, the problem is not trivial especially when exploring the interactions between human actor, object and scene (collectively as human-context) to boost video action detectors. The difficulty originates from the aspect that reliable relations in a video should depend on not only short-term human-context relation in the present clip but also the temporal dynamics distilled over a long-range span of the video. This motivates us to capture both short-term and long-term relations in a video. In this paper, we present a new Long Short-Term Relation Networks, dubbed as LSTR, that novelly aggregates and propagates relation to augment features for video action detection. Technically, Region Proposal Networks (RPN) is remoulded to first produce 3D bounding boxes, i.e., tubelets, in each video clip. LSTR then models short-term human-context interactions within each clip through spatio-temporal attention mechanism and reasons long-term temporal dynamics across video clips via Graph Convolutional Networks (GCN) in a cascaded manner. Extensive experiments are conducted on four benchmark datasets, and superior results are reported when comparing to state-of-the-art methods.
\end{abstract}

%
%


%

%
\maketitle

\section{Introduction}

\begin{figure}[!t]
\centering {\includegraphics[width=0.50\textwidth]{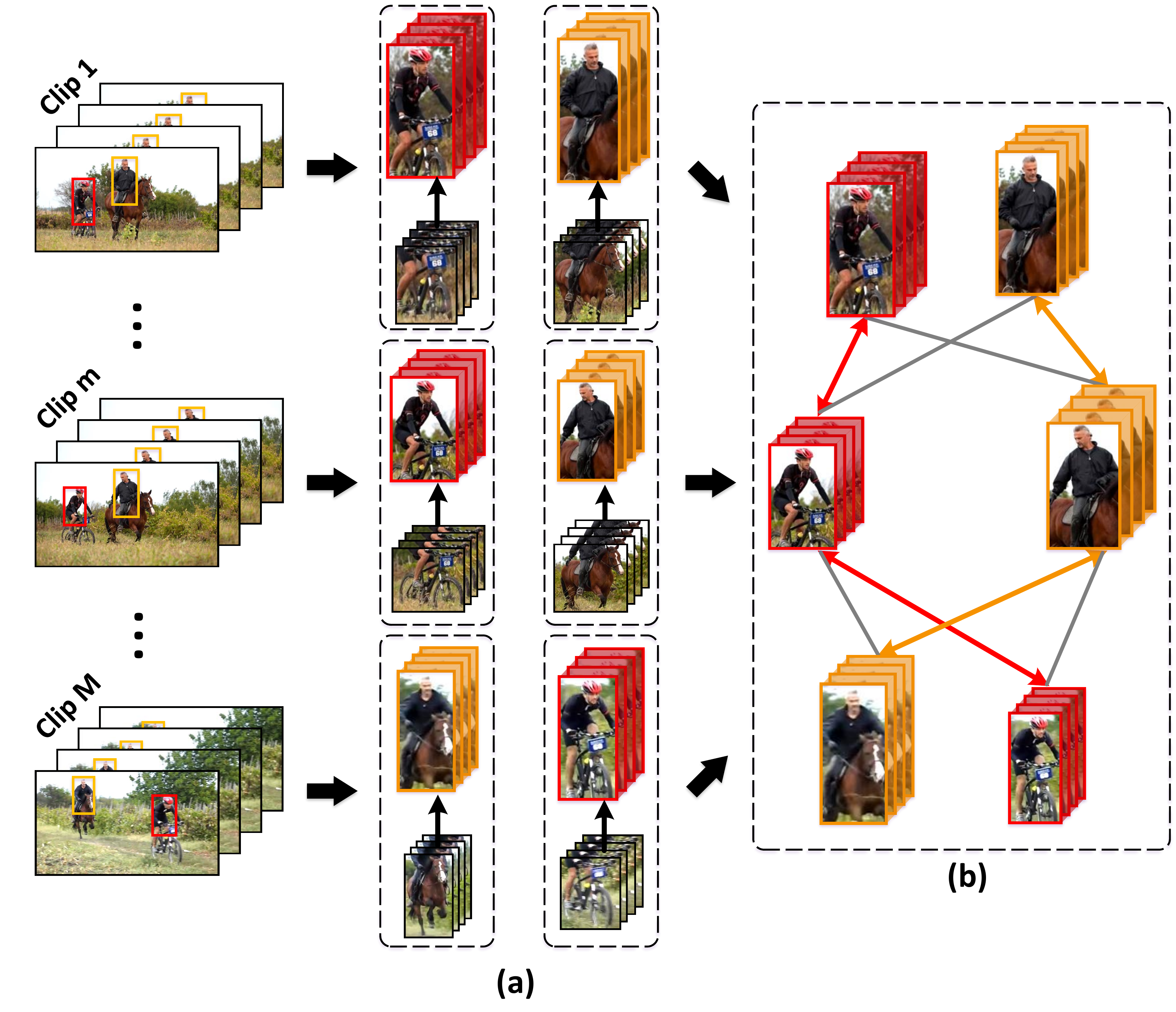}}
\vspace{-0.22in}
\caption{\small Modeling long short-term relation by (a) aggregating short-term interactions within each video clip and (b) propagating long-term temporal dynamics across clips in a cascaded manner.}
\label{fig:intro}
\vspace{-0.18in}
\end{figure}

Convolutional Neural Networks (CNN) have driven video understanding technologies to reach new state-of-the-arts. As one of the most fundamental tasks, object detection in videos has attracted a surge of research interests. The recent methods have proceeded along two dimensions: box-level association \cite{feichtenhofer2017detect,han2016seq,kang2017object,kang2018t} by delving into the association across bounding boxes from consecutive frames, and feature aggregation \cite{wang2018fully,xiao2018video,zhu2018towards,zhu2017flow} via improving per-frame features by aggregation of nearby features. In a further step to localize and recognize actions in videos, video action detection explores spatio-temporal coherence to boost detectors. The research in this area generally follows the way of either performing object detectors on individual frame/motion vectors plus tracking in between or capitalizing on Recurrent Neural Networks and 3D Convolutions to model temporal information and produce action volumes. Regardless of these different recipes for enhancing video action detection, a common issue not fully studied is the exploitation of relations between human, object and scene, i.e., human and surrounding context, which is well believed to be helpful for action~detection.

Visual relations characterize the interactions or relative positions between human actors, relevant objects and scenes. In the literature, there have been strong evidences on the use of visual relation to support various vision tasks, e.g., recognition \cite{wang2018videos}, object detection \cite{hu2018relation} and captioning \cite{yao2018exploring}. One representative work that employs relation for video action detection is introduced in \cite{sun2018actor}. The basic idea is to measure pairwise relation between human actor and global context in each frame through executing 1$\times$1 convolution on the concatenation of actor and global scene features. The method verifies the merit on modeling human-context relation to augment actor features and eventually enhance video action detection. Nevertheless, the mining of relation is implemented spatially on individual frames only and does not involve any temporal information, not to mention the modeling of long-term relation over large temporal extent. Moreover, different actors are treated equally in this case, making it difficult to learn finer relation specific to each actor. To alleviate these issues, we argue that two key ingredients should be taken into account. One is to extend the 2D region proposals in a frame to 3D cubes, i.e., tubelets, in a short video clip. In view that 3D tubelets are rich in spatio-temporal content, the interactions estimated on tubelet level are more comprehensive. More importantly, the perceptions of different human actor tubelets are independent to the global context. As such, we dynamically predict the relevant context for a specific human tubelet on the fly to aggregate human-context relation. We name this in-clip relation as short-term relation, which is illustrated in Figure \ref{fig:intro}(a). The other is to leverage supportive context from long-range temporal dynamics. Following the philosophy that whatever is happening in the present is related to what happened in the past, the propagation between tubelets across video clips encodes long-term temporal context and does benefit action detection. Figure \ref{fig:intro}(b) shows the interaction across temporal clips, which is named as long-term relation.

By consolidating the idea of modeling both short-term and long-term relation in a video, we novelly present Long Short-Term Relation Networks (LSTR) for boosting video action detection. Specifically, Tubelet Proposal Networks (TPN) remoulds Region Proposal Networks (RPN) by extending 2D anchor boxes in RPN to 3D anchor tubelets, and is first exploited to produce human actor tubelets in all video clips. The goal of our LSTR is then to augment the feature of each actor tubelet by aggregating human-context relation within the video clip and propagating long-range temporal relation across video clips. LSTR employs a two-stage reasoning structure, which models short-term relation and long-term relation in a cascaded manner. In the stage of mining short-term relation, LSTR capitalizes on adaptive convolution on each actor tubelet to dynamically predict the spatio-temporal attention map, which indicates the essential degree of the global context to this tubelet. Through 3D attention pooling on the attention map, human-context relation is encoded into context feature, which augments the feature of human tubelet. Next, in the stage of capturing long-term relation, LSTR builds a relation graph with undirected edges on human tubelets extracted from all video clips. The vertex represents each human tubelet and the edge denotes the relation measured on both visual similarity and geometrical overlap in between. Graph Convolutional Networks (GCN) are utilized to enrich the feature of human tubelet by propagating the relation in the graph. The upgraded relation-aware feature of each tubelet is finally exploited for tubelet classification and linking.

The main contribution of this work is the proposal of the use of human-context relation and long-term temporal dynamics for enhancing features of human tubelets and eventually boosting video action detection. This leads to the elegant views of how to produce spatio-temporal tubelets and localize the essential context specifically of each human tubelet, and how to nicely leverage both short-term human-context relation and long-range temporal relation for action detection, which are problems not yet fully~understood.

\section{Related Work}

\textbf{Action recognition} is a fundamental computer vision task and has evolved rapidly in recent years. Early approaches usually rely on hand-crafted features, which detect spatio-temporal interest points and then describe these points with local representations \cite{wang2011action,wang2013action}. Inspired by the advances of Convolutional Neural Networks (CNN) in image classification \cite{krizhevsky2012imagenet}, recent works have attempted to design effective deep architectures for action recognition \cite{karpathy2014large,simonyan2014two,tran2015learning,li2018unified,qiu2017deep}. In \cite{simonyan2014two}, the famous Two-Stream architecture is devised by applying two CNNs separately on visual frames and stacked optical flows.
However, this approach only captures the appearance and local motion feature, which fails to model the long-term temporal dynamics in the video.
To overcome this limitation, Tran \emph{et al.} \cite{tran2015learning} propose a widely adopted 3D CNN (C3D) for learning video representation over 16-frame video clips in the context of large-scale supervised video dataset. Compared to 2D ConvNets, C3D holds much more parameters and is difficult to obtain good convergence.
Consequently, I3D \cite{carreira2017quo} further takes the advantage of ImageNet pretraining by inflating 2D ConvNets into 3D. P3D \cite{qiu2017learning} decouples a 3D convolution filter to a 2D spatial convolution filter and a 1D temporal convolution filter. In this paper, we imitate the design of P3D in the backbone network, but our measure of relation could be readily integrated into any advanced networks.

\textbf{Action detection} aims to spatio-temporally localize a recognized action within a video. Most existing methods \cite{gkioxari2015finding,weinzaepfel2015learning,peng2016multi,saha2016deep,singh2017online} are developed on the basis of 2D object detectors. The typical process is to detect human actions in each frame separately and then link these bounding boxes to create spatio-temporal tubelets. However, inferring actions from single frame could be ambiguous. To address this issue, ACT \cite{kalogeiton2017action} takes a sequence of frames as input and stacks the features from subsequent frames to predict scores. T-CNN \cite{hou2017tube} exploits 3D convolution to estimate short tubes. RTPR \cite{li2018recurrent} applies an LSTM on top of the tubelet features to model long-term temporal dynamics. \cite{gu2018ava} and \cite{wu2019long} propose to utilize I3D as feature extractor, which takes longer video clips as input, but on the assumption that actions appear in the same location across frames. Though the aforementioned methods do take into account temporal information, they overlook spatio-temporal contextual relations between human actors, objects, and the surrounding scene. We demonstrate that endowing the state-of-the-art action detection framework with the leverage of such context relations could provide considerable~improvements.

\begin{figure*}[!tb]
\centering {\includegraphics[width=0.98\textwidth]{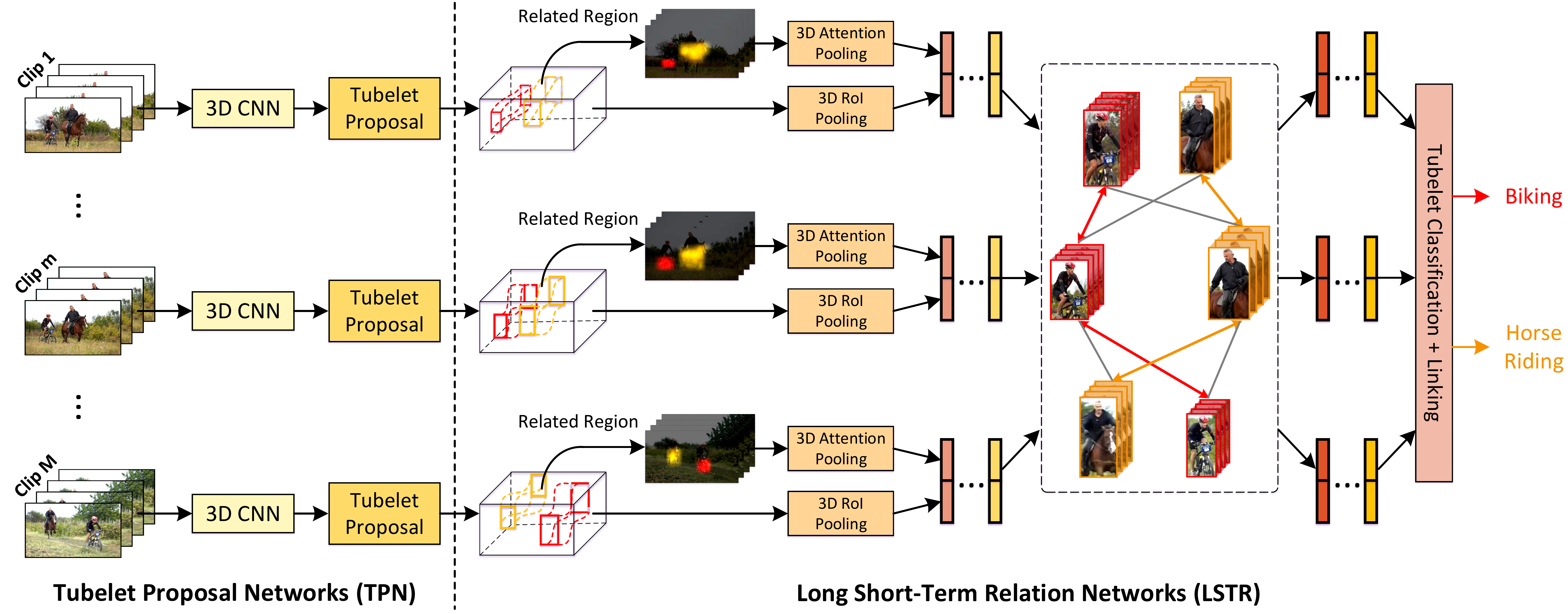}}
\vspace{-0.12in}
\caption{\small An overview of our video action detection framework. Given an input video, a set of fixed-length video clips is produced by evenly dividing the video. The clips are first fed into Tubelet Proposal Networks (TPN) to generate tubelets in parallel.
Our Long Short-Term Relation Networks (LSTR) then augments the feature of each tubelet by aggregating human-context relation within the video clip and propagating long-range temporal relation across video clips in a cascaded manner. Technically, the short-term relation between each actor tubelet and the surrounding context within video clip is unearthed through dynamically predicting a spatio-temporal attention map of this actor tubelet by adaptive convolution. In long-term relation mining, LSTR builds a graph on human tubelets extracted from all video clips and Graph Convolutional Networks (GCN) are utilized to further enrich the feature of human tubelet by propagating the relation in the graph. The upgraded relation-aware feature of each tubelet is finally exploited for tubelet classification and the tubelets across video clips are linked as a video-level sequence spanning over the whole video.}
\label{fig:framework}
\vspace{-0.18in}
\end{figure*}

\textbf{Visual relation modeling} has been proven to be helpful for various tasks including object detection \cite{yao2010modeling,deng2019rdn} and image captioning \cite{yao2018exploring,yao2019hip}. For action recognition, a lot of efforts have been made on modeling pairwise human-object and object-object relationships \cite{ma2018attend,gkioxari2018detecting,qi2018learning}. However, the reasoning of relations in these works is only performed on static images and requires full supervision of actor, action, and objects.
In \cite{wang2018videos}, Wang \emph{et al.} propose to represent videos as space-time region graphs and use Graph Convolutional Networks (GCNs) \cite{kipf2017semi} to capture relations between objects. One limitation of this approach is that objects are detected by an object detector pre-trained on extra training data with a closed vocabulary. In contrast to \cite{wang2018videos}, we build a category-agnostic relation module to detect any context that is highly related to the action in a weakly-supervised manner, without the need of object detectors or extra annotations. The most closely related work is \cite{sun2018actor}, which treats each location in the feature map of the image as an object proxy and computes actor-object relation feature maps via an additional convolutional layer. Ours is different from \cite{sun2018actor} in the way that we adopt an adversarial erasing approach to localize relevant context and devise adaptive convolution on each actor tubelet to dynamically model the human-context relation. Moreover, \cite{sun2018actor} only takes spatial relation into consideration, while our approach integrates both spatial and temporal relations into an unified framework.

\section{Video Action Detection}
\label{sec:method}
The aim of spatio-temporal action detection is to localize and recognize all actors' actions in the videos, which could involve huge diversities on action categories, person poses, and interactions between human or objects.
To cover these aspects, we propose to explicitly build the short-term relation between human and relevant context within each video clip and the long-term relation between tubelets across clips.
An overview of our framework is illustrated in Figure \ref{fig:framework}.
Given an input video, a set of fixed-length video clips \{$S_1,S_2,\cdots,S_M$\} is produced by evenly dividing the video.
The clips are first fed into Tubelet Proposal Networks (TPN) in parallel to generate 3D cuboids (tubelets) of the target actions with actionness scores.
The Long Short-Term Relation Networks (LSTR) is then proposed to augment the feature of each actor tubelet by aggregating human-context relation within video clip and propagating long-range temporal relation across video clips in a cascaded manner.
Specifically, to model the short-term relations between the target actors and their surrounding context (e.g., objects, actors, scenes) in each tubelet, a spatio-temporal attention module is devised to generate an attention map that indicates the correlations in between.
Through 3D Attention Pooling, these relations are encoded into the context feature, which is concatenated with the human feature to form a tubelet representation.
Furthermore, the obtained tubelet representations of multiple tubelets across clips are utilized to model long-term person behaviors through a relation graph.
We perform reasoning on the graph via Graph Convolutional Networks (GCN) to enrich the feature of human tubelet by propagating relations in the graph.
Finally, the upgraded relation-aware feature of each tubelet is employed for classifying actions and linking the tubelets as sequences over the whole video.

\subsection{Tubelet Proposal Networks}
The Tubelet Proposal Networks (TPN) aims to generate the action tubelet candidates for classification.
Considering that action detection task often involves the human-centric actions, we concentrate on human tubelets in TPN.
Most of the state-of-the-art action detection methods \cite{sun2018actor, wu2019long, girdhar2019video} rely on 2D Region Proposal Networks (RPN) \cite{ren2015faster} to generate proposals in each frame independently, but ignore the temporal continuity of videos, which may result in inaccurate detection.
As such, our proposed TPN remoulds RPN by extending 2D anchor boxes to 3D anchor cuboids, that fully utilizes the temporal context for clip-level proposal generation.
TPN takes clips as input and outputs tubelets, i.e., sequences of bounding boxes with associated scores.
The generated 3D tubelets then contain rich and comprehensive information of spatio-temporal content.
Such scheme could be implemented with various state-of-the-art 3D CNN.
In view that P3D \cite{qiu2017learning} provides an economic and flexible structure, we integrate the design of P3D into our TPN, which is more fit for large scale video datasets.

\textbf{Network Architecture.} TPN takes the input of a $T$-frame (typically $16$) clip and first extracts its feature representation via several convolutional and max pooling layers.
Rather than directly applying 3D filters to the clips, we adopt a workaround that treats each 3D filter as a combination of a 2D spatial filter and a 1D temporal filter.
Such framework could not only incorporate temporal context from neighboring frames, but also leverage powerful architecture designs in image domain.
Denote the size of 2D filters as $k \times k$ and 3D filters as $l \times k \times k$, where $l$ is the temporal length and $k$ is the spatial size.
In practice, we use the pre-trained 2D VGG16 network \cite{simonyan2015very} to initialize our TPN.
As described above, we replace each $k \times k$ convolutional layer with a $1 \times k \times k$ convolutional layer followed by a $l \times 1 \times 1$ convolutional layer, where the $1 \times k \times k$ kernel is copied from VGG16.
It is worth noting that we do not utilize any temporal downsampling layers (either temporal pooling or time-strided convolutions) throughout the network.
Our feature maps thus always have $T$ frames along the temporal dimension, preserving the temporal fidelity as much as possible.
The tubelet proposals are generated on the output feature maps of the last convolutional layer.
Specifically, the feature maps are fed into two sibling fully-connected layers, one for regressing tubelet locations and the other for estimating actionness scores.
For regression, we extend the 2D anchor boxes in RPN to 3D anchor cuboids by initializing the spatial extent (i.e., the size and position) of the anchor to be fixed over the $T$ frames.
The regression layer then outputs $4T$ coordinate predictions ($4$ for each of the $T$ frames) for each anchor cuboid. Note that although all boxes in a tubelet are regressed jointly, their locations are different across frames in the clip.
For classification, the layer simply outputs a score for each anchor cuboid indicating whether it is an action or not.

\textbf{Training Objective.} During training, only the clips in which all frames contain the ground-truth actions are considered.
We denote by $\mathcal{A}$ the set of anchor cuboids and by $\hat{\mathcal{A}}$ the ground-truth tubelets.
The overlap between tubelets is defined by averaging the Intersection over Union (IoU) between boxes over $T$ frames.
We assign a positive label to the anchor cuboid $a_i \in \mathcal{A}$ if: (a) $a_i$ has an IoU overlap higher than $0.5$ with any ground-truth tubelets $\hat{a_i} \in \hat{\mathcal{A}}$, or (b) $a_i$ is with the highest IoU overlap with a ground-truth tubelet $\hat{a_i} \in \hat{\mathcal{A}}$. With these definitions, the overall training objective $\mathcal{L}$ is formulated as
\begin{equation}\label{Eq:loss}
\begin{aligned}
\mathcal{L} = \frac{1}{N}\sum_{i} \mathcal{L}_{cls}(p_i, \hat{p_i})+\lambda\frac{1}{N_{reg}}\sum_i \hat{p_i} \mathcal{L}_{reg}(a_i, \hat{a_i}).
\end{aligned}
\end{equation}
Here, $i$ is the index of anchor cuboid $a_i$ in a mini-batch and $p_i$ is the predicted actionness score of $a_i$. The ground-truth label $\hat{p_i}$ is 1 if $a_i$ is assigned a positive label, otherwise 0. $a_i$ is a vector indicating the $4T$ parameterized coordinates of the predicted tubelet, and $\hat{a_i}$ is the corresponding vector of the ground-truth tubelet. $N$ is the mini-batch size, $N_{reg}$ is the number of positive proposals, and $\lambda$ is the parameter for balancing classification and regression. The classification loss $\mathcal{L}_{cls}$ is the standard softmax loss and the regression loss $\mathcal{L}_{reg}$ is the smooth $L_1$ loss \cite{girshick2015fast}.

\subsection{Long Short-Term Relation Networks}

Once the action proposals are produced, conventional action detection methods directly extract the RoI pooled features for classification.
However, unlike object detection which mostly depends on the visual appearance, action detection involves much more variation and can hardly be achieved via only appearance.
For example, to differentiate the actions of ``dunk'' and ``rebound'' in basketball, the model is excepted to not only capture the co-occurrence of the player and the ball, but also reason out the interactions or relationships between them.
To achieve this, we propose to model both the short-term relation within each clip and the long-term relation across multiple clips to augment the feature of each tubelet.

\begin{figure}[!t]
\centering {\includegraphics[width=0.49\textwidth]{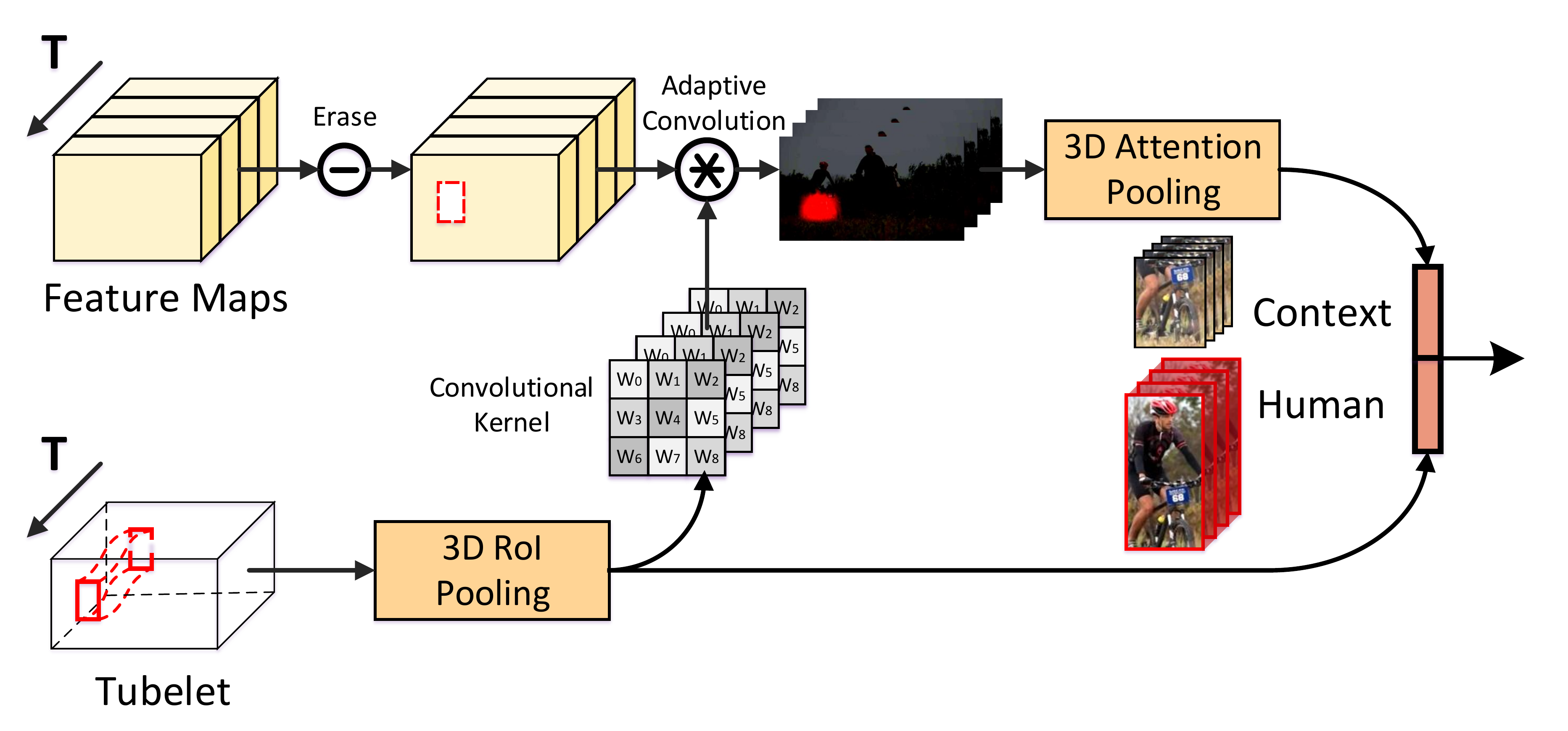}}
\vspace{-0.2in}
\caption{\small The measure of short-term relation. An adaptive convolution kernel is learnt on human tubelet feature.
By convolving the kernel over an adversarially erased feature map, a tubelet-specific spatial-temporal attention map is dynamically predicted to capture complementary regions with high relation to the tubelet.
The attention map indicates the essential degree of the global context to this tubelet.
With 3D Attention Pooling, the context feature is obtained and then concatenated with human feature.}
\label{fig:attention}
\vspace{-0.2in}
\end{figure}

\subsubsection{\textbf{Short-Term Relation}} \label{sec:short}
The short-term relation within clip presents the interactions between actors and their surroundings (including other actors, objects, and scenes), i.e., human-context relation.
A spatio-temporal attention module is devised to model and incorporate such information into tubelet representation, as illustrated in Figure \ref{fig:attention}.
We exploit adaptive convolution to dynamically predict the tubelet-specific spatio-temporal attention map, which indicates the relevance degree of the global context to this tubelet.
The context feature is then generated through 3D Attention Pooling on the attention map, which encodes human-context~relation.

Our proposed spatio-temporal attention module takes as inputs of the clip feature representation $\bm{F} \in \mathbb{R}^{T \times H \times W \times C}$ and the tubelet proposals from TPN, where $H, W, C$ denotes the height, width, channel number of the feature.
For each human tubelet $h$, we apply 3D RoI Pooling to $\bm{F}$ and obtain a human representation $\bm{F}^h$ of size $T \times 7 \times 7 \times C$.
Different from the conventional practice of directly using $\bm{F}^h$ for action classification, we aim to generate a set of weights that indicate the relations between the actor and every elements in the clip.
On the assumption that each individual feature cell in the feature map $\bm{F}$ represents an element, this set of weights, i.e., attention map $\bm{A}^h \in \mathbb{R}^{T \times H \times W}$ could be learned on $\bm{F}^h$ and $\bm{F}$.
One typical solution is to imitate the self-attention mechanism \cite{vaswani2017attention,girdhar2019video} and directly map the concatenation of $\bm{F}^h$ and the elements in $\bm{F}$ to the attention weights via a fully-connected layer.
However, such simple linear projection has difficulty in learning specific weights for each tubelet due to the variation among them.
Consequently, we propose to learn an adaptive convolution kernel for each tubelet, which is further utilized to dynamically generate the tubelet-specific attention map.
Formally, the adaptive convolution kernel $\bm{K}^h\in \mathbb{R}^{T \times 3 \times 3}$ for the tubelet could be estimated on actor's visual feature $\bm{F}^h$ as
\begin{equation} \label{Eq:kernel}
\bm{K}^h = g_{\theta}(\bm{F}^h),
\end{equation}
where $g_{\theta}$ is the transformation function parameterized by $\theta$. In practice, we implement this function by a fully-connected layer. The kernel $\bm{K}^h$ provides the specific essentials for capturing human-context relation based on human feature. The spatio-temporal attention map is thus obtained by convolving the kernel $\bm{K}^h$ over the clip representation $\bm{F}$, which can be formulated as
\begin{equation} \label{Eq:conv}
\bm{A}^h = \sigma(\bm{K}^h \ast \bm{F}) ,
\end{equation}
where $\sigma$ is the sigmoid function and $(\ast)$ denotes 3D convolution.

The main challenge in the above solution is that the obtained attention map has a high overlap with the tubelet itself, while the other active regions are very small and sparse.
This is not surprising because the classification networks are inclined to identify patterns from the most discriminative parts (human cuboids in our case) for recognition \cite{zhou2016learning}.
To eliminate the influence of target tubelet, we adopt an adversarial erasing scheme to discover new and complementary context by erasing the region of target tubelet from the feature map $\bm{F}$.
The obtained map $\bm{F}^e$, rather than $\bm{F}$, is utilized in Eq. (\ref{Eq:conv}).
With such adversarial learning, the attention map is forced to capture complementary context regions that are highly related to the target tubelet.

Given the attention map $\bm{A}^h$, we compute the feature representation $\widehat{f}^h$ of those relevant elements, i.e., context, by 3D Attention Pooling, which multiplies $\bm{F}$ by $\bm{A}^h$ via element-wise product and then aggregates the obtained map over all locations,
\begin{equation}\label{Eq:4}
\widehat{f}^h = \sum_{t=1}^T\sum_{i=1}^H\sum_{j=1}^W \bm{A}^h_{t,i,j} \odot \bm{F}_{t,i,j} .
\end{equation}
To generate the final tubelet feature $f$, we first project the human representation $\bm{F}^h$ to a low-dimensional embedding $f^h$, and then concatenate it with the context representation $\widehat{f}^h$:
\begin{equation}\label{Eq:5}
f = [f^h;\widehat{f}^h] .
\end{equation}
The feature $f$ is a new representation of the actor and able to model the interactions with the surrounding spatio-temporal regions.

\subsubsection{\textbf{Long-Term Relation}}
In addition to the short-term relation between actors and context within each clip, we expect to further capitalize on long-range dependencies between correlated tubelets from neighboring clips.
To achieve this, we construct a relation graph to model the long-term temporal dynamics across clips by propagating the relation in the graph.

Suppose $N_m$ tubelets $\{h_1,h_2,\ldots,h_{N_m}\}$ are generated from the $m$-th clip. We stack the extracted short-term tubelet features as $ \mathbf{X}_m=[f_1,f_2,\ldots,$ $f_{N_m}]$, where $\mathbf{X}_m \in \mathbb{R}^{N_m \times d}$ and $d$ is the dimension of $f$.
Our target is to mine the long-term relation for each tubelet and leverage the relation to enrich the tubelet features, obtaining an updated version of $\mathbf{X}_m$.
To this end, we carry it out within a long temporal window centered at current clip ($m$-th) with radius $w$, whose representation is $\mathbf{X}=[\mathbf{X}_{m-w},\ldots,\mathbf{X}_{m},\ldots,\mathbf{X}_{m+w}] \in \mathbb{R}^{N \times d}$, $N=\sum_{m'=m-w}^{m+w} N_{m'}$.
Defining each tubelet as a graph vertex, a bipartite relation graph $\mathbf{G}$ between $\mathbf{X}_m$ and $\mathbf{X}$ is constructed by connecting vertexes from the two sets.
Typically, strong connections are expected between pairs which are semantically and spatially related.
Specifically, the edge will have a high weight if the two tubelets are: (a) the same actor or both related to target actions, or (b) highly overlapped in geometrical space.

Formally, the weight of edge $e_{ij}$ between the $i$-th tubelet $h_i$ from $\mathbf{X}_m$ and the $j$-th tubelet $h_j$ from $\mathbf{X}$ can be represented as
\begin{equation}\label{Eq:edge}
e_{ij}={\phi(f_i)}^\top\phi(f_j)+\gamma\cdot iou(h_i,h_j),
\end{equation}
where $f_i$ and $f_j$ are the features of $h_i$ and $h_j$, $\phi(\cdot)$ represents the feature transformation, $iou(\cdot)$ is the IoU overlap ratio of two tubelets, and $\gamma$ is a scalar parameter.
In practice, we have $\phi(f)=\mathbf{w}f+\mathbf{b}$, where $\mathbf{w}$ is of size $d \times d$ and $\mathbf{b}$ is the bias vector.
Such transformation is able to learn not only the relations between different states of the same actor, but also relations between different actors.
As indicated in Eq. (\ref{Eq:edge}), the edge score $e_{ij}$ consists of two factors, i.e., the semantic similarity and the geometrical overlap. The first factor, ${\phi(f_i)}^\top\phi(f_j)$, measures the semantic similarity between tubelets. The second factor, $iou(h_i,h_j)$, captures their geometrical correlations.

After computing the edge score matrix in Eq. (\ref{Eq:edge}), we perform normalization on each row of the matrix so that the sum of all the edge values connected to one tubelet is $1$. Particularly, we adopt the softmax function and obtain the graph by
\begin{equation}\label{Eq:norm}
\mathbf{G}_{ij}=\frac{exp(e_{ij})}{\sum_{j=1}^{N}exp(e_{ij})} .
\end{equation}
We then exploit the Graph Convolutional Networks (GCN) \cite{kipf2017semi} for performing reasoning on the graph $\mathbf{G}$.
Different from conventional CNN which performs convolution on a local regular grid, GCN operates on graphs and induces features of nodes based on their neighborhoods.
Therefore, the relations between tubelets are propagated through the graph while performing the graph convolutions, which provide complementary characteristic for each tubelet.
We represent the graph convolution as
\begin{equation}\label{Eq:GCN}
\mathbf{Z}_m=\mathbf{GXW},
\end{equation}
where $\mathbf{G}$ is the relation graph introduced in Eq. (\ref{Eq:norm}) with $N_m \times N$ dimensions, $\mathbf{X}$ is the feature set of the long temporal window, and $\mathbf{W}$ is the weight parameter matrix of size $d \times d$.
The generated $\mathbf{Z}_m\in \mathbb{R}^{N_m \times d}$ is our final augmented tubelet representation and could be regarded as the weighted aggregation of~$\mathbf{X}$, which encodes both short-term and long-term relations.
We feed $\mathbf{Z}_m$ into the classifier for action prediction, and then perform the linking across clips by utilizing the linking algorithm in \cite{li2018recurrent} to obtain the video-level tubelet spanning over the whole video.

\section{Implementations}
We implement our video action detection framework on Caffe toolbox \cite{jia2014caffe} and exploit VGG16 model \cite{simonyan2015very} pre-trained on ImageNet \cite{russakovsky2015imagenet} to initialize TPN as in \cite{qiu2017learning}. When training TPN, each video is divided into $16$-frame clips of resolution $320 \times 240$ with stride $1$. The scalar parameter $\lambda$ in Eq. (\ref{Eq:loss}) is set to $1$. Since the output tubelet proposals of TPN are highly overlapped, non-maximum suppression (NMS) is adopted to reduce redundancy and merge the tubelets based on their actionness scores.
We fix the IoU threshold in NMS to $0.7$ and utilize the top $300$ tubelet proposals for action classification.
In the mining of long-term relation, the radius of temporal window is set as $w=4$, which is a good trade-off between performance and GPU memory demand.
If the number of clips in a video is less than $2w+1$, we pad zeros after the last frame. The scalar parameter $\gamma$ in Eq.(\ref{Eq:edge}) is set to 1. The final action classifier is a softmax layer for single-label prediction or a sigmoid layer for multi-label prediction. An extra dropout layer is further added with dropout ratio 0.5 before the softmax/sigmoid layer. Following \cite{peng2016multi,singh2017online,kalogeiton2017action,li2018recurrent}, we also exploit a two-stream pipeline for utilizing multiple modalities, where the RGB frame and the stacked optical flow ``image'' are considered. To fuse the detection results, late fusion scheme is taken to average the classification scores.

We train our model using stochastic gradient descent (SGD) with a mini-batch size of 16 clips on 8 GPUs, \emph{i.e.}, 2 clips per GPU. We warm-up the learning rate from $0.0001$ to $0.001$ in the first $0.3$ epochs using linear annealing and then employ cosine learning rate decay \cite{loshchilov2017sgdr}. The whole training procedure stops at 10 epochs.
The momentum and weight decay are set to 0.9 and 0.0001, respectively.

\section{Experiments}
\subsection{Datasets and Evaluation Metrics}
We evaluate our LSTR on four popular action detection datasets: UCF-Sports \cite{rodriguez2008action}, J-HMDB \cite{jhuang2013towards}, UCF-101 \cite{soomro2012ucf101}, and AVA \cite{gu2018ava}.

\textbf{UCF-Sports} includes 150 short videos from 10 sport classes. All videos are well trimmed and contain spatio-temporal annotations. We follow the standard split in \cite{lan2011discriminative} for training and testing.

\textbf{J-HMDB} consists of 928 videos of 21 different actions. Human silhouettes are annotated for all frames and the ground-truth bounding boxes are inferred from the silhouettes. Three training/testing splits are officially provided by the dataset organizers. Following \cite{kalogeiton2017action,li2018recurrent}, we conduct the elaborated study of each ingredient on the first split, and report the average performances over three splits in the comparison with the state-of-the-art approaches.

\textbf{UCF-101} is widely adopted for action recognition. For action detection task, a subset of 24 action classes and $3,207$ videos are provided with spatio-temporal annotations. Different from UCF-Sports and J-HMDB in which videos are truncated to actions, videos in UCF-101 are untrimmed. Similar to \cite{peng2016multi,saha2016deep}, we only report the results on the first training/testing split.

\textbf{AVA} v2.1 consists of 211K training, 57K validation, and 117K testing clips taken at one clip per second from 430 different movies. Unlike the above datasets in which annotations are given for all frames, every person only in the middle frame of each clip is localized in a bounding box and labeled with (possibly multiple) actions from the vocabulary of AVA. There are 80 different atomic actions in the vocabulary. Following the official settings in \cite{gu2018ava}, we conduct the experiments on a subset of 60 classes and each class contains at least 25 validation examples. For comparisons with the state-of-the-art methods, we additionally submit our method to online ActivityNet Challenge server and report the performances on testing~set.

\textbf{Evaluation metrics.} We adopt frame-level and video-level mean Average Precision (frame-mAP and video-mAP) for evaluation. A detection is positive if its IoU with a ground-truth box or tubelet is greater than a threshold $\delta$ and the action is also predicted correctly. Specifically, for UCF-Sports, J-HMDB, and UCF-101, we follow the standard evaluation scheme \cite{peng2016multi} to measure video-mAP. For AVA, we follow \cite{gu2018ava} to compute frame-mAP. The evaluation is performed at IoU threshold $\delta=0.2$ on UCF-101 and $\delta=0.5$ on UCF-Sports, J-HMDB and AVA, unless otherwise stated.

\begin{table}[]
\caption{\small Performance comparisons of modeling human-context relation by different ways on UCF-Sports, J-HMDB, and UCF-101 datasets. The video clips are in the form of both RGB frames and optical flow images.}
\begin{tabular}{l|cc|cc|cc}
\hline
\multirow{2}*{\textbf{Method}}  & \multicolumn{2}{c|}{\textbf{UCF-Sports}} & \multicolumn{2}{c|}{\textbf{J-HMDB}} & \multicolumn{2}{c}{\textbf{UCF-101}} \\
           & RGB            & Flow           & RGB          & Flow         & RGB          & Flow         \\ \hline
Concat     & 86.4           & 95.0           & 60.8         & 78.0         & 59.8         & 73.4         \\
Concat+Att & 86.9           & 95.6           & 60.8         & 78.1         & 60.4         & 73.8         \\
\textbf{STR}       & \textbf{87.9}           & \textbf{96.8}           & \textbf{62.0}         & \textbf{79.0}         & \textbf{61.9}         & \textbf{75.3}         \\ \hline
\end{tabular}
\label{table:shortterm}
\vspace{-0.1in}
\end{table}

\begin{figure*}[!tb]
\centering {\includegraphics[width=0.98\textwidth]{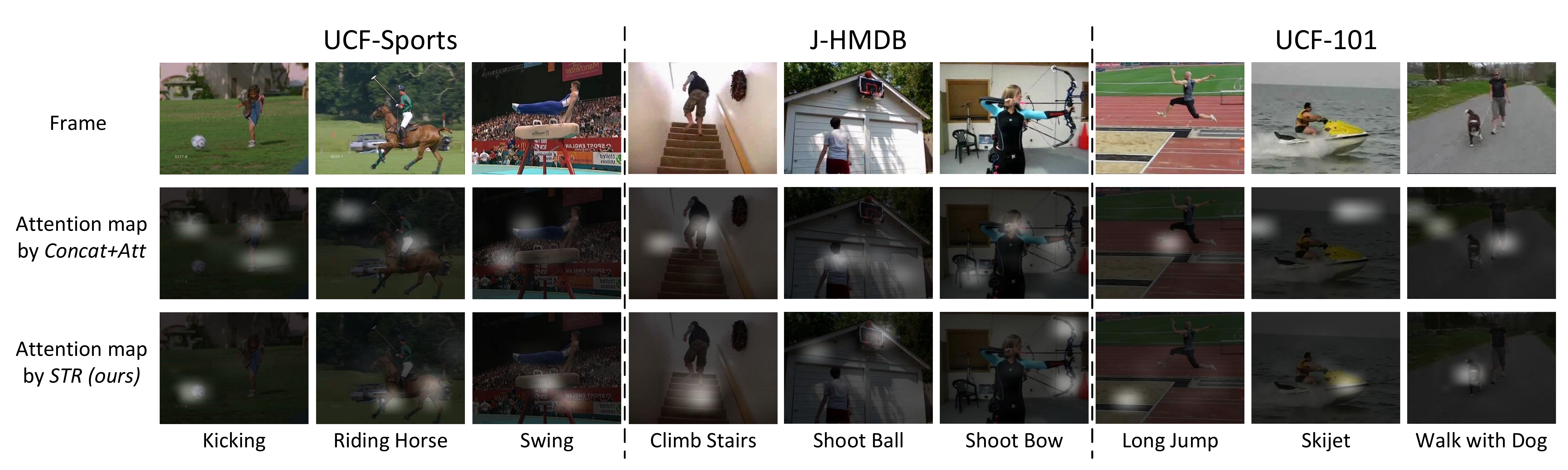}}
\vspace{-0.15in}
\caption{\small An illustration of learnt attention maps on nine video clips by Concat+Att and our STR. Top row: one sampled frame from each video clip; second row: attention map learnt by Concat+Att; third row: attention map learnt by our STR, where the brightness indicates the strength of focus. (better viewed in color)}
\label{fig:shortterm}
\vspace{-0.15in}
\end{figure*}

\subsection{Evaluation on Short-Term Relation}
We first evaluate the measure of short-term relation (STR) in LSTR and compare our STR with the way of Concatenation (Concat) in \cite{sun2018actor} and Concatenation plus Attention (Concat+Att) in \cite{ulutan2018actor}. Concat aims to estimate pairwise relation between human actor and global context irrespective of the localization of essential context. For each actor tubelet, a fixed-length feature vector is extracted via 3D RoI Pooling followed by average pooling over all spatio-temporal locations. Such feature vector is tiled and concatenated to features extracted from global context at every spatio-temporal location. The method of Concat then executes $1 \times 1$ convolution on the combined feature map $\bm{F}^c$ to produce pair-wise relation feature, which is finally averaged over spatio-temporal locations to obtain final context feature. Instead of directly utilizing $\bm{F}^c$ in Concat, Concat+Att learns attention map by performing $1 \times 1$ convolution on $\bm{F}^c$ and then multiplies attention map with global context to generate context feature.

Table \ref{table:shortterm} summarizes the video-mAP performances when exploiting different measures of human-context relation on UCF-Sports, J-HMDB, and UCF-101, respectively. The results across three datasets consistently indicate that our measure of STR leads to performance boost against the other two methods on the input video clips in the form of RGB frames or optical flow images. Our STR and Concat+Att focusing on relevant context by attention mechanism also exhibit better performances than Concat which capitalizes on the whole context. The result basically indicates the advantage of modeling functional relation on essential context. Though both STR and Concat+Att learn attention map through Convolution, they are fundamentally different in the way that Concat+Att exploits the same conv filter for all the tubelets, while our STR dynamically predicts the parameters of conv filter adapted to each specific tubelet. As indicated by our results, allowing tubelet-specific conv filter can lead to better performance gain. Figure \ref{fig:shortterm} illustrates the learnt attention maps on nine video clips by Concat+Att and our STR. An observation is that Concat+Att concentrates on the essential elements in general, which often include actor tubelets themselves. Our STR, in comparison, is benefited from adversarial erasing operation and only focuses on the complementary context.
Take the first/last action of kicking/walk with dog as examples, Concat+Att often pinpoints to actor cuboids, while our STR only captures complementary context, i.e., football/dog in two cases. Moreover, STR also eliminates unrelated context, e.g., lawn in the last action.

\subsection{Evaluation on Long-Term Relation}
Next, we turn to examine our design of long-term relation (LTR) based on the augmented feature of actor tubelet with context feature in STR. As such, STR is regarded as a base point, which directly does classification on the output feature of STR. In view that the relations between tubelets are built on two factors of visual similarity and geometrical overlap in LTR, we validate the impact of each towards the performance. The run of \emph{Similarity} or \emph{Overlap} takes only one factor of visual similarity or geometrical overlap into account when constructing the relation graph.

\begin{table}[]
\caption{\small Performance comparisons of building long-term relation on different factors. The video clips are in the form of both RGB frames and optical flow images.}
\vspace{-0.1in}
\begin{tabular}{l|cc|cc|cc}
\hline
\multirow{2}*{\textbf{Method}}   & \multicolumn{2}{c|}{\textbf{UCF-Sports}} & \multicolumn{2}{c|}{\textbf{J-HMDB}} & \multicolumn{2}{c}{\textbf{UCF-101}} \\
           & RGB            & Flow           & RGB          & Flow         & RGB          & Flow         \\ \hline
STR        & 87.9           & 96.8           & 62.0         & 79.0         & 61.9         & 75.3         \\
\emph{Similarity} & 88.4           & 97.3           & 62.9         & 79.7         & 63.0         & 76.8         \\
\emph{Overlap}    & 88.1           & 97.0           & 62.5         & 79.3         & 62.5         & 75.9         \\
\textbf{LSTR}       & \textbf{89.1}           & \textbf{97.7}           & \textbf{63.1}         & \textbf{80.2}         & \textbf{63.9}         & \textbf{77.6}         \\ \hline
\end{tabular}
\label{table:longterm}
\vspace{-0.1in}
\end{table}

\begin{figure}[!t]
\centering {\includegraphics[width=0.49\textwidth]{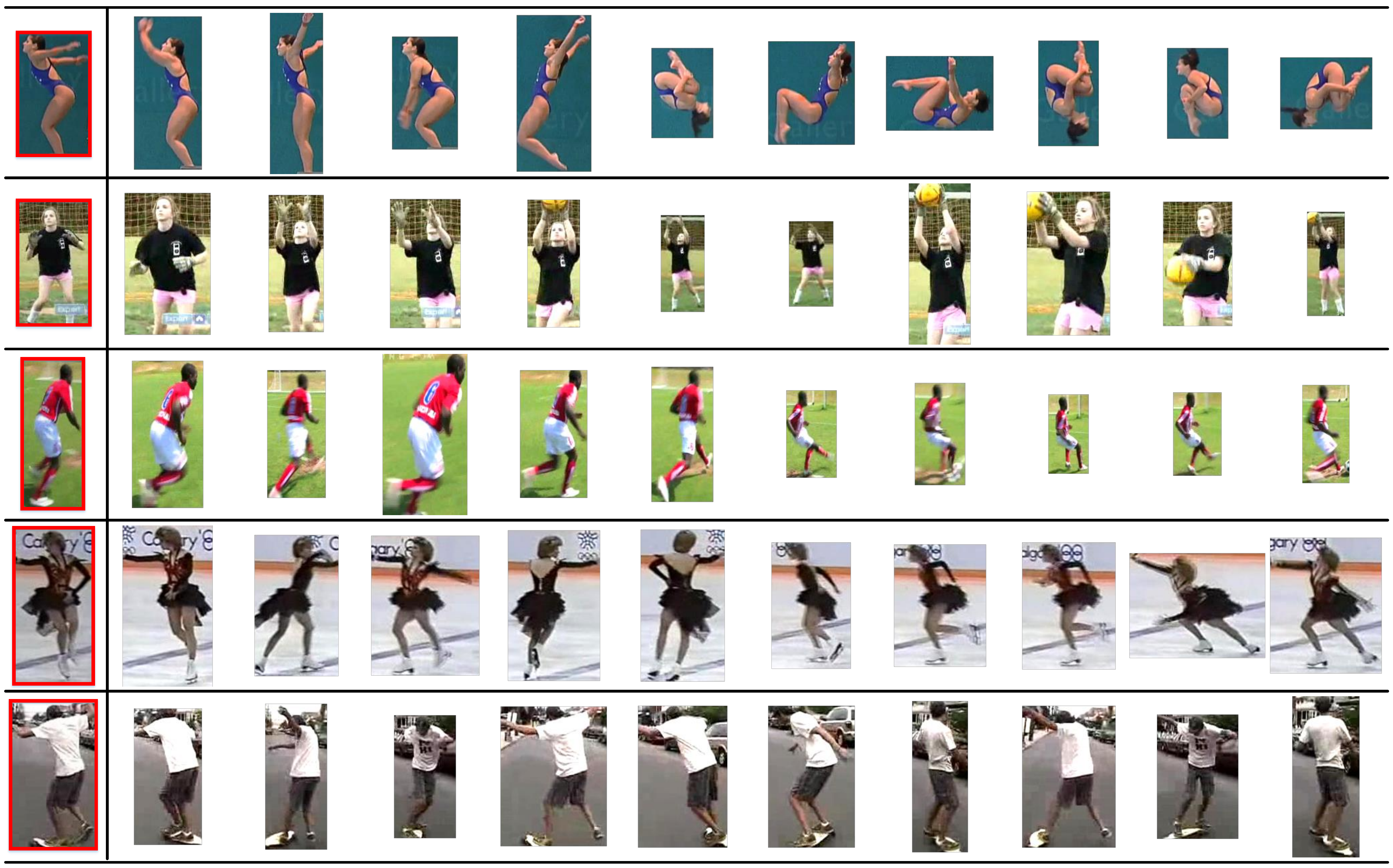}}
\vspace{-0.2in}
\caption{\small Examples showing the top-10 most related tubelets from the relation graph in response to five given tubelets. In each row, the first one in a red rectangle represents the given tubelet.}
\label{fig:longterm}
\vspace{-0.2in}
\end{figure}

Table \ref{table:longterm} details the performance comparisons. Building the long-term relation on either visual similarity or geometrical overlap constantly improves the performances on three datasets. The results verify the idea of reasoning long-term temporal dynamics across video clips for action detection. When jointly considering the two factors, LSTR further boosts up the performances. That basically demonstrate the complementarity between the two factors in modeling long-term relation. Note that because UCF-101 has longer temporal context ($\sim$ 200 frames per video) than UCF-Sports and J-HMDB ($\sim$ 60 and 30 frames per video), it is not surprise that LSTR leads to larger performance gains on UCF-101 (2.0\%/2.3\%) than UCF-Sports (1.2\%/0.9\%) and J-HMDB (1.1\%/1.2\%). Figure \ref{fig:longterm} showcases the top-10 most related tubelets in our graph in response to five given tubelets from UCF-Sports, J-HMDB, and UCF-101. The bounding box in the middle of the tubelet is shown to represent each tubelet. As shown in the Figure, all the top-10 tubelets describe the identical human actor over long-range temporal span of the video and provide more supportive context to infer the action.

\subsection{Ablation Study}
Here, we study how each design in LSTR influences the overall performance. Faster R-CNN simply executes RPN on frame level and exploits 2D RoI Pooling feature for action classification. \textbf{TPN} remoulds RPN to generate 3D tubelets on clip level and employs 3D RoI Pooling feature for classification. \textbf{Att} leverages the idea of adaptive convolution to predict spatio-temporal attention on the fly for each tubelet in our STR. \textbf{Erase} further integrates the adversarial erasing operation into STR. \textbf{LTR} is our long-term relation module.

\begin{table*}[]
\caption{\small Performance contribution of each design in LSTR.}
\vspace{-0.1in}
\begin{tabular}{l|cccc|ccc|ccc|ccc}
\hline
\multirow{2}*{\textbf{Method}}  & \multirow{2}*{\textbf{TPN}}  & \multirow{2}*{\textbf{Att}} & \multirow{2}*{\textbf{Erase}} & \multirow{2}*{\textbf{LTR}} & \multicolumn{3}{c|}{\textbf{UCF-Sports}} & \multicolumn{3}{c|}{\textbf{J-HMDB}} & \multicolumn{3}{c}{\textbf{UCF-101}} \\
            &     &     &       &     & RGB      & Flow     & Fusion    & RGB     & Flow   & Fusion   & RGB     & Flow   & Fusion   \\ \hline
Faster R-CNN &     &     &       &     & 84.2     & 94.5     & 95.1      & 57.3    & 75.0   & 76.8     & 57.4    & 71.2   & 72.9     \\
+TPN        &$\surd$  &     &       &     & 86.3     & 95.1     & 96.4      & 59.9    & 77.2   & 79.2     & 60.0    & 73.2   & 74.8     \\
+Att        &$\surd$  &$\surd$  &       &     & 87.2     & 96.1     & 97.3      & 61.4    & 78.5   & 80.7     & 61.2    & 74.5   & 76.2     \\
+Erase      &$\surd$  &$\surd$  &$\surd$   &     & 87.9     & 96.8     & 98.2      & 62.0    & 79.0   & 81.2     & 61.9    & 75.3   & 76.8     \\ \hline
\textbf{LSTR}        &$\surd$  &$\surd$  &$\surd$   &$\surd$   & \textbf{89.1}     & \textbf{97.7}     & \textbf{98.8}      & \textbf{63.1}    & \textbf{80.2}   & \textbf{82.1}     & \textbf{63.9}    & \textbf{77.6}   & \textbf{78.7}     \\ \hline
\end{tabular}
\label{table:ablation}
\end{table*}

Table \ref{table:ablation} shows the video-mAP performances by involving each design in LSTR. Take the input of RGB frames in a video as an example, directly performing Faster R-CNN on frame level achieves 84.2\%, 57.3\% and 57.4\% of video-mAP on UCF-Sports, J-HMDB, and UCF-101. TPN is a straightforward way to encode spatio-temporal information in the stage of tubelet proposal. In our case, TPN successfully leads to a video-mAP increase of 2.1\%, 2.6\% and 2.6\% on three datasets, respectively. Att and Erase are two specific designs in STR. The performance gains of each on three datasets are 0.9\%/1.5\%/1.2\% and 0.7\%/0.6\%/0.7\%. In total, our STR produces a video-mAP boost of 1.6\%/2.1\%/1.9\%. The results verify the merit of STR. With a further contribution of LTR on the integration of long-range temporal dynamics, LSTR finally reaches 89.1\%, 63.1\% and 63.9\% of video-mAP. Similar trends are observed on the input of optical flow images or the fusion of the two types of inputs.

\subsection{Comparison with State-of-the-art}
We compare with several state-of-the-art techniques on four benchmarks. For fair comparisons, we also utilize ResNet101 \cite{he2016deep} as the backbone in our TPN. Following \cite{peng2016multi,singh2017online,kalogeiton2017action,li2018recurrent}, we report the performance of LSTR on the late fusion of RGB images and optical flow images inputs. Table \ref{table:state-of-the-art1} summarizes video-mAP performances on UCF-Sports, J-HMDB (3 splits) and UCF-101 datasets with different IoU thresholds $\delta$. In particular, Faster R-CNN based approaches \cite{peng2016multi,saha2016deep} utilizing RPN and RoI pooling outperform R-CNN based methods \cite{gkioxari2015finding,weinzaepfel2015learning}. T-CNN \cite{hou2017tube} and ACT \cite{kalogeiton2017action} improve \cite{peng2016multi,saha2016deep} by modeling short-term temporal information. RTPR \cite{li2018recurrent} exploring long-term temporal dynamics with LSTM further boosts up the performance. \cite{sun2018actor} leads to video-mAP gain by modeling the relation between human and global context, but still yields inferior performance to our LSTR. The results again demonstrate the advantage of modeling both short-term and long-term relation in LSTR.

\begin{table}[tb]\small
\centering
\caption{\small Video-mAP comparisons on UCF-Sports, J-HMDB, and UCF-101. The performances on J-HMDB are averaged over 3 splits.}
\label{table:state-of-the-art1}
\begin{tabular}{l|cc|cc|cc}
\hline
\multirow{2}{*}{\textbf{Method}}                       & \multicolumn{2}{c|}{\textbf{UCF-Sports}} & \multicolumn{2}{c|}{\textbf{J-HMDB}}  & \multicolumn{2}{c}{\textbf{UCF-101}}       \\ \cline{2-7}
                                                       & 0.2            & 0.5                     & 0.2   & 0.5    & 0.2   & 0.5               \\ \hline
Gkioxari \emph{et al.} \cite{gkioxari2015finding}       & -              & 75.8                    & -     & 53.3   & -     & -\\
Weinzaepfe \emph{et al.} \cite{weinzaepfel2015learning} & -              & 90.5                    & 63.1  & 60.7   & 46.8  & -\\
Saha \emph{et al.} \cite{saha2016deep}                  & -              & -                       & 72.6  & 71.5   & 66.8  & 35.9\\
Peng \emph{et al.} \cite{peng2016multi}\footnotemark[1] & 94.8           & 94.7                    & 74.3  & 73.1   & 72.9  & -\\
Singh \emph{et al.} \cite{singh2017online}              & -              & -                       & 73.8  & 72.0   & 73.5  & 46.3\\
Kalogeiton \emph{et al.} \cite{kalogeiton2017action}    & 92.7           & 92.7                    & 74.2  & 73.7   & 77.2  & 51.4\\
Hou \emph{et al.} \cite{hou2017tube}\footnotemark[2]    & 95.2           & 95.2                    & 78.4  & 76.9   & 73.1  & -\\
Yang \emph{et al.} \cite{yang2017spatio}                & -              & -                       & -     & -      & 73.5  & 37.8\\
He \emph{et al.} \cite{he2018generic}                   & 96.0           & 95.7                    & 79.7  & 77.0   & 71.7  & -   \\
Li \emph{et al.} \cite{li2018recurrent}                 & 98.6           & 98.6                    & 82.7  & 81.3   & 77.9  & -\\
Gu \emph{et al.} \cite{gu2018ava}                       & -              & -                       & -     & 78.6   & -     & 59.9   \\
Sun \emph{et al.} \cite{sun2018actor}                   & -              & -                       & -     & 80.1   & -     & -   \\ \hline
\textbf{LSTR}                                                   &                &                         &       &        &       & \\
~~-{w/ TPN$_{VGG16}$}                                         & 98.8           & 98.8                    & 83.1  & 81.7   & 78.7  & 53.8\\
~~-{w/ TPN$_{ResNet101}$}                                     & \textbf{98.9}  & \textbf{98.9}           & \textbf{86.9} & \textbf{85.5}  & \textbf{83.0}  & \textbf{64.4}\\ \hline
\end{tabular}
\vspace{-0.1in}
\end{table}
\footnotetext[1]{Updated results from \url{https://hal.inria.fr/hal-01349107/file/eccv16-pxj-v3.pdf}}
\footnotetext[2]{Updated results from \url{https://arxiv.org/pdf/1712.01111.pdf}}

Table \ref{table:state-of-the-art2} details frame-mAP on both AVA validation and test sets. The frame-mAP of our LSTR achieves 27.2\% which is a new record on validation set, making the absolute improvement over the best competitor \cite{wu2019long} by 1.4\%. Moreover, we also submit our LSTR to online ActivityNet Challenge server and evaluate LSTR on test set. Again, LSTR outperforms other runs.

\subsection{Radius $w$ of Temporal Window}
In LSTR, the radius $w$ of temporal window in long-term relation mining is a free parameter. In the previous experiments, the radius is empirically set to 4. Furthermore, we conduct experiments to test the impact of this parameter towards detection performance. The frame-mAP on AVA validation set fluctuates within the range of 0.003, when $w$ are set from 3 to 6. The best performance of 27.2\% is attained when $w=4$. Once the radius is large than 3, the performance is less affected with the change of temporal range, which eases the selection of radius in our LSTR practically.

\begin{table}[]
\caption{\small Frame-mAP comparisons on AVA validation and test sets.}
\label{table:state-of-the-art2}
\begin{tabular}{l|cc}
\hline
\textbf{Method}                                                      & \textbf{validation}  & \textbf{test} \\ \hline
Gu \emph{et al.} \cite{gu2018ava}                             & 15.8 & -    \\
Sun \emph{et al.} \cite{sun2018actor}                         & 17.4 & -    \\
Ulutan \emph{et al.} \cite{ulutan2018actor}                   & 22.5 & -    \\
Jiang \emph{et al.} \cite{jiang2018human}                     & 25.6 & 21.1 \\
Wu \emph{et al.} \cite{wu2019long}                            & 25.8 & 24.8 \\
Girdhar \emph{et al.} \cite{girdhar2019video}                 & 24.4 & 24.3 \\ \hline
\textbf{LSTR}                                                          &      &      \\
~~-{w/ TPN$_{VGG16}$}                                         & 14.4 & -    \\
~~-{w/ TPN$_{ResNet101}$}                                     & \textbf{27.2} & \textbf{25.3}    \\ \hline
\end{tabular}
\vspace{-0.1in}
\end{table}

\section{Conclusions}
We have presented Long Short-Term Relation Networks (LSTR) architecture, which models both short-term and long-term relation to boost video action detection. Particularly, we study the problem from the viewpoint of employing human-context relation within each video clip and leveraging supportive context from long-range temporal dynamics. To verify our claim, we utilize Tubelet Proposal Networks to generate 3D actor tubelets in all video clips. For each actor tubelet, LSTR dynamically predicts the spatio-temporal attention map on the fly via adaptive convolution to indicate the essential context and measures human-context relation on the attention map. Such short-term relation is encoded into context feature to augment the feature of tubelet. Moreover, LSTR builds a graph on all the actor tubelets and capitalizes on Graph Convolutions to propagate the long-term temporal relation over the graph and further enrich tubelet feature. Extensive experiments conducted on four benchmark datasets validate our proposal and analysis. More remarkably, we achieve new state-of-the-art performances on AVA dataset.

\newpage
\balance

\end{document}